\documentclass{article}
\usepackage[preprint]{neurips_2025}
\usepackage[utf8]{inputenc} 
\usepackage[T1]{fontenc}    
\usepackage{hyperref}       
\usepackage{url}            
\usepackage{booktabs}       
\usepackage{amsfonts}       
\usepackage{nicefrac}       
\usepackage{microtype}      
\usepackage{xcolor}         
\usepackage{algorithmic}
\usepackage{multirow}
\usepackage{graphicx}
\usepackage[table]{xcolor}
\usepackage{amsmath}
\usepackage{subcaption}
\usepackage{dsfont}
\usepackage{cleveref}
\usepackage{algorithm2e}
\usepackage{enumitem}
\newcommand{\myfirstpara}[1]{\noindent {\bf #1:}}
\newcommand{\mypara}[1]{\vspace{0.5em} \myfirstpara{#1}}

\definecolor{bestgreen}{RGB}{200,255,200}
\definecolor{secondred}{RGB}{255,220,220}

\def\NC{$\mathcal{NC}$}

\title{Reliable Active Learning from Unreliable Labels via Neural Collapse Geometry}
\workshoptitle{Reliable ML from Unreliable Data}

\author{
  Atharv Goel\thanks{Equal contribution.} \\
  IIIT Delhi \\
  \texttt{atharv21027@iiitd.ac.in} \\
  \And
  Sharat Agarwal\footnotemark[1] \\
  IIIT Delhi \\
  \texttt{sharata@iiitd.ac.in} \\
  \And
  Saket Anand \\
  IIIT Delhi \\
  \texttt{anands@iiitd.ac.in} \\
  \AND
  Chetan Arora \\
  IIT Delhi \\
  \texttt{chetan@cse.iitd.ac.in} \\
}

\begin{document}

\maketitle
\begin{abstract}
Active Learning (AL) promises to reduce annotation cost by prioritizing informative samples, yet its reliability is undermined when labels are noisy or when the data distribution shifts. In practice, annotators make mistakes, rare categories are ambiguous, and conventional AL heuristics (uncertainty, diversity) often amplify such errors by repeatedly selecting mislabeled or redundant samples. We propose Reliable Active Learning via Neural Collapse Geometry (NCAL-R), a framework that leverages the emergent geometric regularities of deep networks to counteract unreliable supervision. Our method introduces two complementary signals: (i) a \textbf{Class-Mean Alignment Perturbation score}, which quantifies how candidate samples structurally stabilize or distort inter-class geometry, and (ii) a \textbf{Feature Fluctuation score}, which captures temporal instability of representations across training checkpoints. By combining these signals, NCAL-R prioritizes samples that both preserve class separation and highlight ambiguous regions, mitigating the effect of noisy or redundant labels. Experiments on ImageNet-100 and CIFAR100 show that NCAL-R consistently outperforms standard AL baselines, achieving higher accuracy with fewer labels, improved robustness under synthetic label noise, and stronger generalization to out-of-distribution data. These results suggest that incorporating geometric reliability criteria into acquisition decisions can make Active Learning less brittle to annotation errors and distribution shifts, a key step toward trustworthy deployment in real-world labeling pipelines.
Our code is available at \href{https://github.com/Vision-IIITD/NCAL}{\texttt{https://github.com/Vision-IIITD/NCAL}}.
\end{abstract}
\section{Introduction}
Deep learning depends on large-scale annotations~\cite{lecun2015deep}, but real-world labels are often unreliable. This undermines Active Learning (AL)~\cite{sener2018active}, whose heuristics (uncertainty~\cite{yoo2019learning, gal2017deep}, diversity~\cite{sener2018active, cdal}) can even exacerbate noise by selecting mislabeled, redundant, or ambiguous samples. This leads to inefficient label use, degraded generalization, and poor robustness under distribution shifts \cite{chitta2021training, wang2022caffe}.

Neural Collapse (NC) theory~\cite{papyan2020prevalence} shows that, late in training, features concentrate near class means, which align as a simplex ETF. These regularities provide stability even under imperfect supervision, suggesting that sample selection guided by NC dynamics could improve both efficiency and robustness \cite{haaslinking, neco}.

In this paper, we propose NCAL-R, a Neural Collapse–guided Active Learning framework designed for reliability under noisy or uncertain supervision. By quantifying how candidate samples perturb inter-class alignment and fluctuate across training checkpoints, NCAL-R selects points that both preserve feature structure and expose genuine ambiguities. Our experiments demonstrate improved accuracy with fewer labels, enhanced robustness to synthetic noise, and stronger out-of-distribution generalization.
\section{Methodology}
\label{sec:method}

\mypara{Problem Setting}
We consider a pool-based Active Learning (AL) setting: a small labeled set $\mathcal{L}$ and a large unlabeled set $\mathcal{U}$ across $K$ classes, with a model $f_\theta$ trained on $\mathcal{L} = \bigcup_{c=1}^K \mathcal{L}^c$, where $\mathcal{L}^c$ is the set of labelled samples of class $c$. At each acquisition step $t$, an AL strategy selects a batch $\mathcal{B}_t \subset \mathcal{U}_t$ for annotation, such that $\mathcal{L}_{t+1} = \mathcal{L}_t \cup B_t$. Our goal is to select $\mathcal{B}_t$ such that the learned representation is \emph{robust} to covariate shift and label drift, enabling improved in-distribution accuracy, OOD detection, and novel-class discovery.

\mypara{Acquisition Metrics}
The emergence of ETF structure in class means (Sec \ref{sec:related}) reflects high class separability, and deviations from it may indicate \textit{structurally valuable} samples that when labeled, could improve generalization. \texttt{NCAL} thus computes two complementary scores for each $x \in \mathcal{U}_t$:

\begin{enumerate}[leftmargin=*]
    \item \textbf{Class-Mean Alignment Perturbation (CMAP)}: 
    The generalization error of a classifier is bound by a term involving the KL divergence between the posterior and prior over model parameters, according to Theorem 3.1 of ~\cite{NEURIPS2020_f48c04ff}. This KL term is correlated with the \textit{weight correlation} at each network layer (Corollary 4.4, ~\cite{NEURIPS2020_f48c04ff}). NC allows us to construct a feature space analog of the weight correlation, as samples collapse to their means, which in turn collapse to their respective classifier weights.
    Define \textit{class-mean alignment} as the mean pair-wise cosine similarity of class-means, where $\mu_t^c = \frac{1}{|\mathcal{L}_t^c|} \sum_{x \in \mathcal{L}_t^c} f_\theta(x)$:
    \begin{equation}
        \mathrm{CMA}(\mathcal{L}_t) = \frac{1}{K(K-1)} \sum_{\substack{i,j=1 \\ i \neq j}}^{K} \text{Sim}(\mu_t^i, \mu_t^j)        
    \end{equation}
    If the candidate sample $x$ has penultimate-layer feature $z$ and the model predicts label $\hat{y}(x) = c$, let $\tilde{\mu}_t^c := \frac{|\mathcal{L}_t^c|\mu_t^c+z}{|\mathcal{L}_t^c|+1}$, i.e., the updated class mean computed across $\mathcal{L}_t \cup x$.
    Our objective is to select samples that minimize \textit{CMA}, the feature space analog of weight correlation. To measure the marginal impact of candidate $x$, we define \textit{CMAP} as the difference between \textit{CMA} computed with $\tilde{\mu}^c$ versus computed with $\mu^c$. This expression simplifies to a dot product:

    \begin{align}
        \mathrm{CMAP}(x) &:= \mathrm{CMA}(\mathcal{L}_t \cup x) - \mathrm{CMA}(\mathcal{L}_t) \nonumber \\
        &= \big(\bar{\tilde{\mu}}_t^{c} - \bar{\mu}_t^{c}\big)^\top \big(M_t - \bar{\mu}_t^{c}\big) \label{eq:cmap}
    \end{align}

    where $\bar{h} := \frac{h}{||h||_2}$ represents the unit-norm version of any vector $h$, and $M_t :=\sum_{i=1}^{K}\bar{\mu}_t^i$ is the sum of unit-normalized class means.
    A high CMAP indicates high deviation from CMA, and hence, high perturbation to feature geometry. By training on samples with high perturbation, we \textit{reduce} the CMA, an approximation of the weight correlation, and hence reduce generalization error. The detailed derivation for Equation \ref{eq:cmap} can be found in Appendix \ref{sec:derivation}.
    
    \item \textbf{Feature Fluctuation (FF)}: Given model checkpoints $\{\theta_t\}_{t=T_i}^{T_f}$ where $T_i$ and $T_f$ are the start and end epochs of the terminal phase (TPT), 
    FF measures the variance of predicted logits for $x$ across $\theta_t$. High FF identifies samples with persistent uncertainty, even when most features have stabilized.
\begin{equation}
    \begin{split}
    \mathrm{FF}(x) = \sum_{t=T_i+1}^{T_f} 
       \mathbf{1}\Big[ & \hat{y}_t(x) \neq \hat{y}_{t-1}(x) \Big]
    \end{split}
\end{equation}

\end{enumerate}

\mypara{Combined Acquisition Strategy}
\texttt{NCAL} selects the top-$k$ samples from $\mathcal{U}$ by ranking CMAP and FF separately, standardizing each by their mean and standard deviation, and averaging:
\begin{equation}
\mathrm{Score}(x) = \frac{\mathrm{CMAP}(x) + \mathrm{FF}(x)}{2}
\end{equation}

This yields a batch $\mathcal{B}$ that contains both structurally impactful and prediction-unstable samples, shaping the representation to be both discriminative and adaptable.
\texttt{NCAL} requires no auxiliary networks, pseudo-labeling, or task-specific tuning, and can be applied to any backbone or modality where feature embeddings can be extracted. The pseudocode for this algorithm can be found in Appendix \ref{sec:pseudocode}.
\section{Experiments}
\label{sec:experiments}

\paragraph{Experimental Setup.}
We evaluate NCAL-R on tasks including classification, OOD detection, OOD generalization, and general category discovery. Label drift is tested under the GCD protocol; covariate shift via linear probes on OOD datasets. Unless noted, we use a ResNet-18 backbone, 5\% acquisition per cycle, and compare to Random, CoreSet~\cite{sener2018active}, and CDAL~\cite{cdal}. The active learning setup is detailed in Appendix \ref{sec:training-protocol}.

\paragraph{Evaluation Metrics.}
We report: (i) \textbf{All-class accuracy}: top-1 classification accuracy over both known and novel classes; (ii) \textbf{Novel-class accuracy}: GCD accuracy restricted to novel classes; (iii) \textbf{Known-class accuracy}: classification accuracy on known classes; (iv) \textbf{AUROC} for binary OOD detection between in-distribution and OOD samples.

\begin{figure*}[ht]
    \centering
    \begin{subfigure}[t]{0.32\textwidth}
        \centering
        \includegraphics[width=\linewidth]{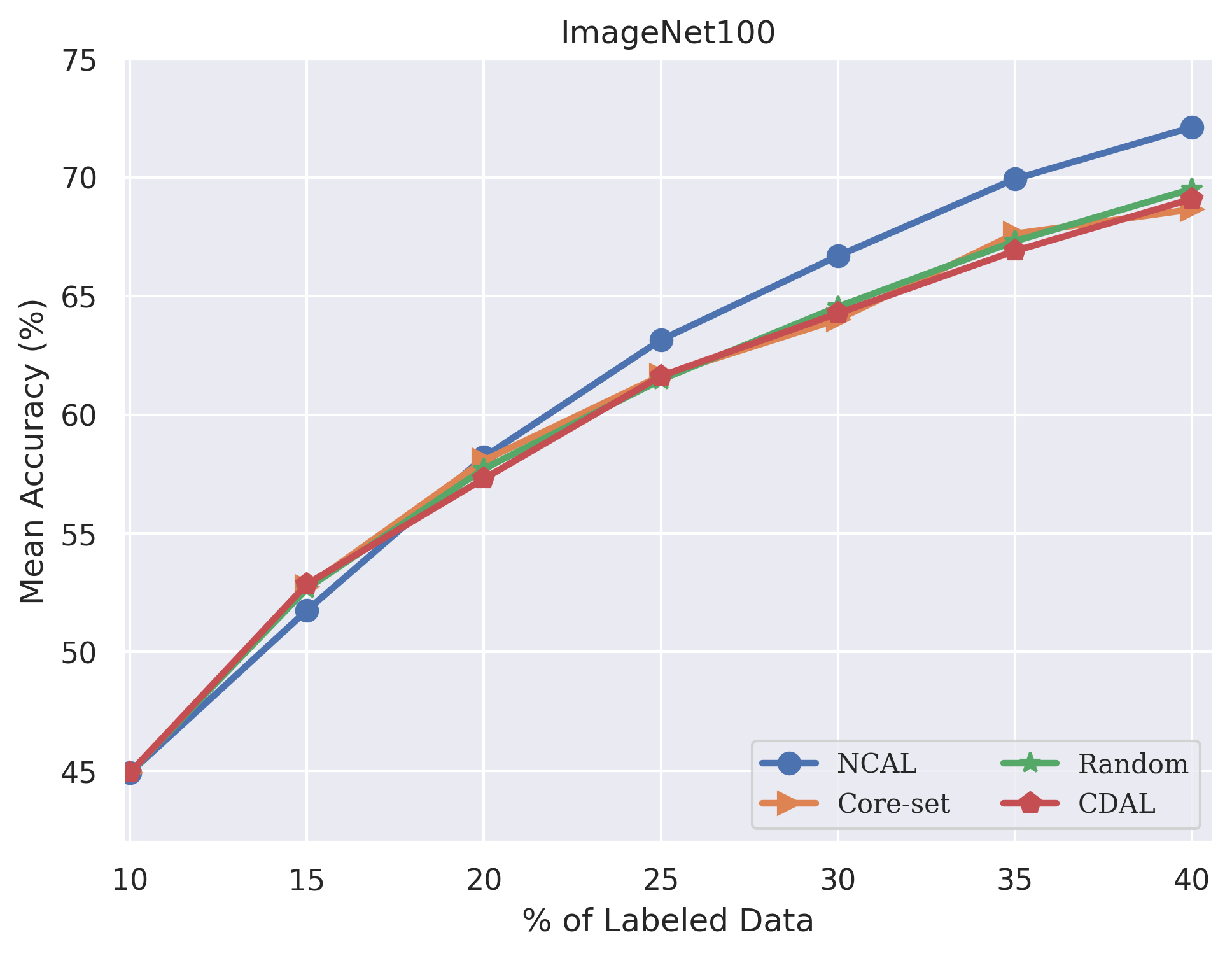}
        
        \label{fig:imagenet100}
    \end{subfigure}
    \hfill
    \begin{subfigure}[t]{0.32\textwidth}
        \centering
        \includegraphics[width=\linewidth]{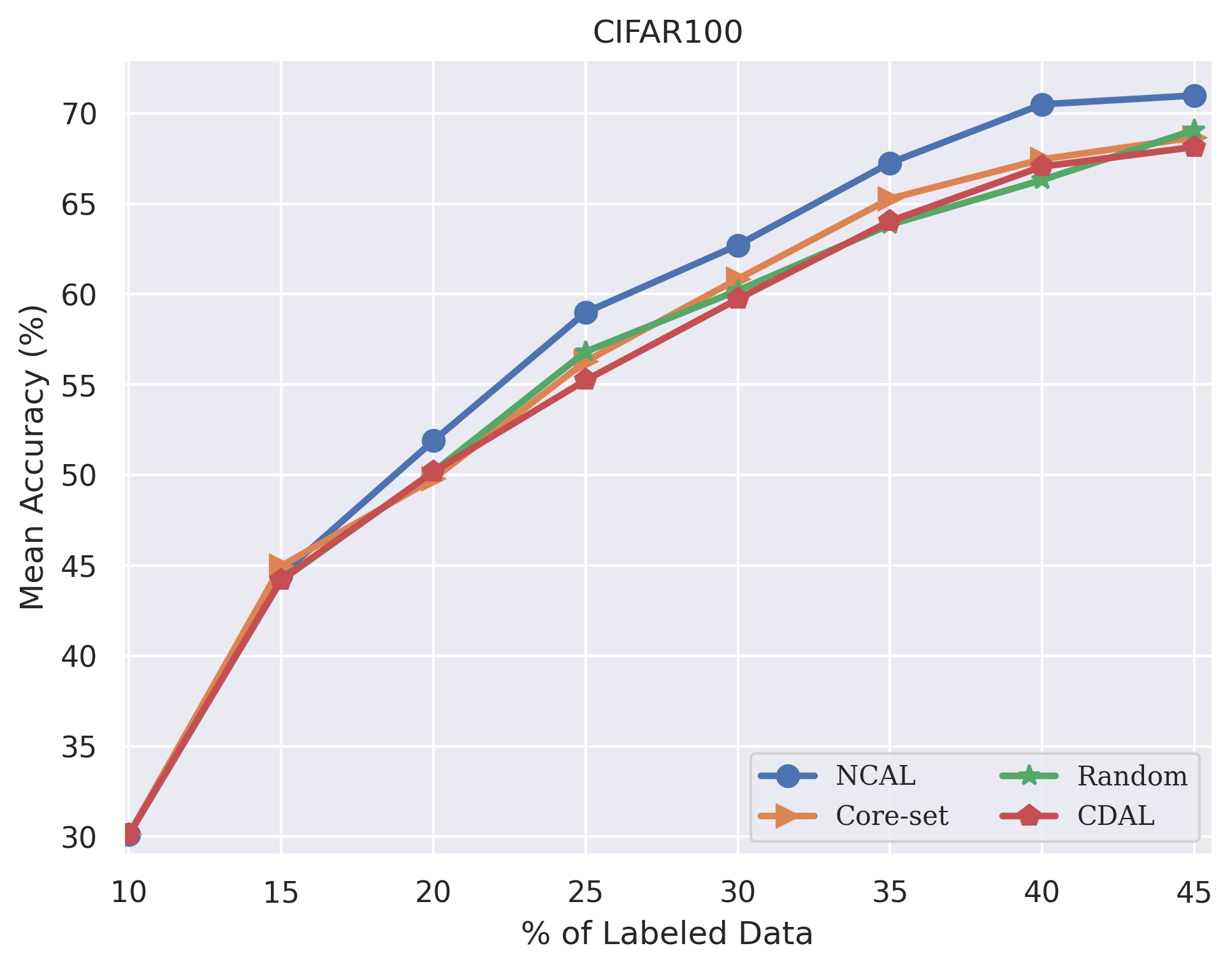}
        
        \label{fig:cifar100}
    \end{subfigure}
    \hfill
    \begin{subfigure}[t]{0.32\textwidth}
        \centering
        \includegraphics[width=\linewidth]{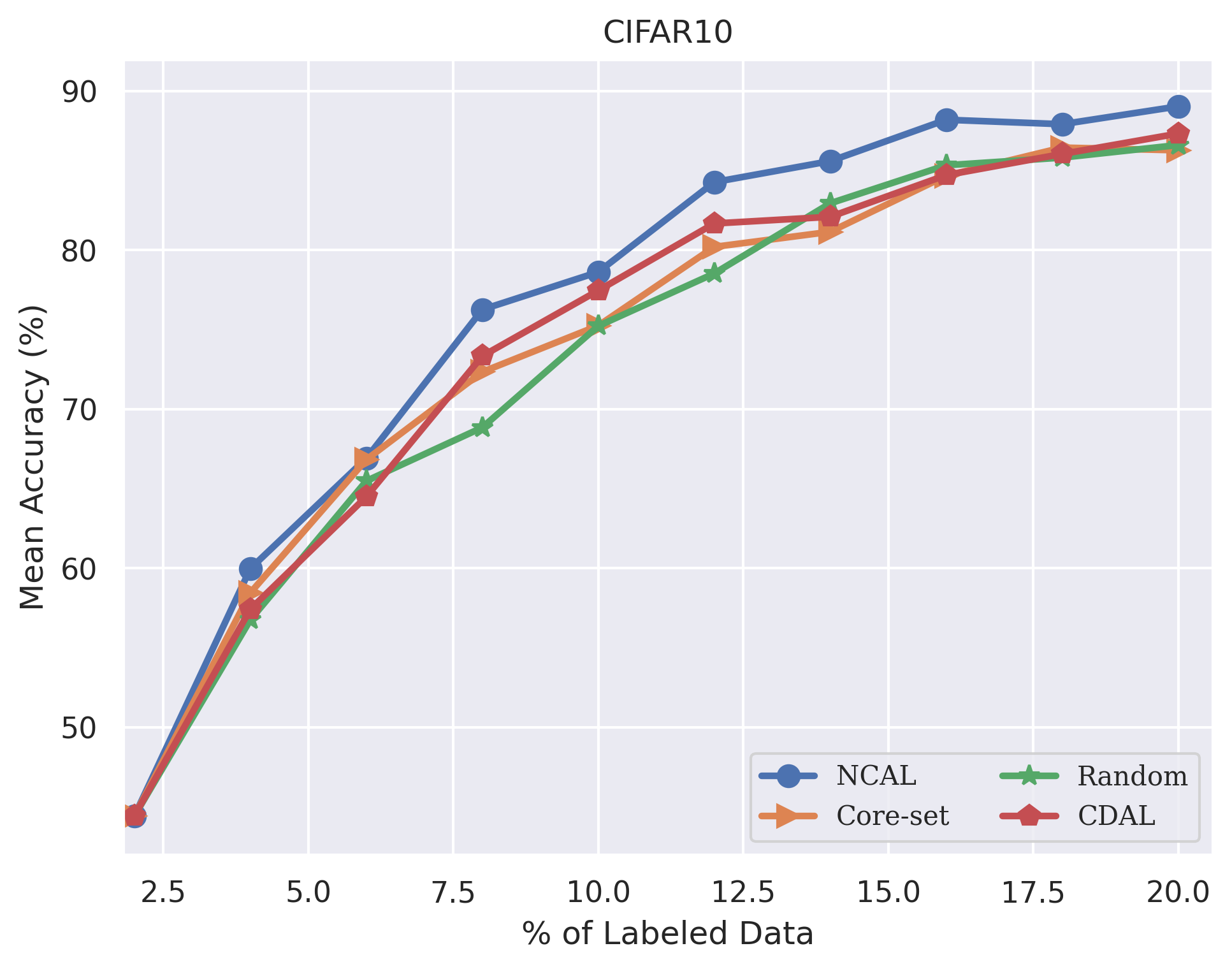}
        
        \label{fig:cifar10}
    \end{subfigure}
    \caption{Comparison of test accuracy across varying label budgets on three benchmark datasets—ImageNet100, CIFAR100, and CIFAR10. \texttt{NCAL}'s good performance even at lower annotation budgets suggests that its Neural Collapse-guided selection promotes more structured and representative feature learning. \textit{(Note: accuracy for 100\% data of ImageNet100, CIFAR100 and CIFAR10 are: 79.16\%, 70.75\% and 90\% respectively. Reported results are average of 3 independent runs.)}}
    \label{fig:AL_exeperiment}
\end{figure*}

\begin{table}[ht]
\centering
\scriptsize
\setlength{\tabcolsep}{4pt}
\begin{tabular}{l|c|c|c|c|c|c|c|c}
\toprule
\textbf{Method} & \textbf{10\%} & \textbf{15\%} & \textbf{20\%} & \textbf{25\%} & \textbf{30\%} & \textbf{35\%} & \textbf{40\%} & \textbf{100\%} \\
\midrule
Random      & \multirow{4}{*}{77.18} & 80.57 & 84.13 & 85.45 & 86.89 & 87.82 & 88.67 & \multirow{4}{*}{93.68} \\
CDAL        &       & 81.78 & 84.28 & 85.9 & 86.34 & 87.98 & 88.92 & \\
Coreset     &       & 81.56 & 83.73 & 85.66 & 87.1 & 88.29 & 88.95 & \\
NCAL        &       & \cellcolor{bestgreen} \textbf{82.49} & \cellcolor{bestgreen} \textbf{85.55} & \cellcolor{bestgreen} \textbf{87.89} & \cellcolor{bestgreen}\textbf{89.15} &  \cellcolor{bestgreen} \textbf{90.53} & \cellcolor{bestgreen}\textbf{91.53} & \\
\bottomrule
\end{tabular}

\caption{AUROC scores for Far-OOD detection on the OpenImage-O dataset trained on ImageNet-100 with varying annotation budgets.}
\label{tab:far_OOD_imagenet100}
\end{table}

\begin{table}[ht]
\centering
\scriptsize
\begin{tabular}{l|cccc}
\toprule
\textbf{Method} & \textbf{All Classes} & \textbf{Old Classes} & \textbf{New Classes} & \textbf{Val Accuracy} \\
\midrule
Random  & 33.20 & 50.34 & 20.35 & 36.20 \\
CDAL    & 33.39 & 49.96 & 20.96 & 36.94 \\
Coreset & 32.23 & 49.98 & 18.92 & 36.44 \\
\textbf{NCAL}    & \cellcolor{bestgreen}\textbf{35.07} & \cellcolor{bestgreen}\textbf{51.95} & \cellcolor{bestgreen}\textbf{23.05} & \cellcolor{bestgreen}\textbf{37.76} \\
\bottomrule
\end{tabular}
\caption{Performance across all, old, and new classes along with validation accuracy.}
\label{tab:gcd_60_40}
\end{table}
\begin{table}[ht]
\centering
\scriptsize
\setlength{\tabcolsep}{4pt}
\begin{tabular}{l|c|c|c|c|c|c|c|c}
\toprule
\textbf{Method} & \textbf{10\%} & \textbf{15\%} & \textbf{20\%} & \textbf{25\%} & \textbf{30\%} & \textbf{35\%} & \textbf{40\%} & \textbf{100\%} \\
\midrule
Random      & \multirow{4}{*}{77.18} & 80.57 & 84.13 & 85.45 & 86.89 & 87.82 & 88.67 & \multirow{4}{*}{93.68} \\
CDAL        &       & 81.78 & 84.28 & 85.9 & 86.34 & 87.98 & 88.92 & \\
Coreset     &       & 81.56 & 83.73 & 85.66 & 87.10 & 88.29 & 88.95 & \\
NCAL        &       & \cellcolor{bestgreen} \textbf{82.49} & \cellcolor{bestgreen} \textbf{85.55} & \cellcolor{bestgreen} \textbf{87.89} & \cellcolor{bestgreen} \textbf{89.15} & \cellcolor{bestgreen} \textbf{90.53} & \cellcolor{bestgreen} \textbf{91.53} & \\
\bottomrule
\end{tabular}
\caption{AUROC scores for Far-OOD detection on the OpenImage-O dataset trained on Imagenet-100 with varying annotation budgets.}
\label{tab:near_OOD_imagenet100}
\end{table}

\begin{table*}[ht]
\centering
\scriptsize
\begin{tabular}{l|cc|ccccccccc}
\toprule
\multirow{2}{*}{} & \multicolumn{2}{c|}{\textbf{Val / Train Acc}}  & \multicolumn{9}{c}{\textbf{OOD Generalization (linear probe val accuracy)}} \\
\cmidrule(r){2-3} \cmidrule(r){4-12}
 & Val (\%) & Train &  ImgNet-R & CIFAR100 & Flowers & NINCO & CUB & Aircraft & Pets & STL & Avg \\
\midrule
Random  & 69.51 & 96.44 & 18.06 & 41.64 & 58.69 & 64.23 & 37.84 & 15.26 & 42.34 & 68.67 & 46.95 \\
CDAL    & 69.09 & 96.55 & 17.56 & 41.98 & 58.13 & 65.87 & 38.53 & 15.03 & 42.65 & 68.27 & 47.21 \\
Coreset & 68.65 & 96.42 & 16.93 & 42.02 & 57.96 & 65.11 & 37.86 & 15.15 & 42.22 & 68.68 & 47.00 \\
NCAL    & \cellcolor{bestgreen}\textbf{72.11} & \cellcolor{bestgreen}\textbf{95.22} & \cellcolor{bestgreen}\textbf{19.27} & \cellcolor{bestgreen}\textbf{43.78} &\cellcolor{bestgreen} \textbf{60.87} & \cellcolor{bestgreen}\textbf{67.66} &\cellcolor{bestgreen} \textbf{40.01} &\cellcolor{bestgreen} \textbf{15.38} &\cellcolor{bestgreen} \textbf{44.70} & \cellcolor{bestgreen}\textbf{70.49} &\cellcolor{bestgreen} \textbf{48.98} \\ \hline
100\%   & 79.16 & 95.27 & 20.01 & 45.31 & 61.77 & 69.90 & 42.29 & 19.08 & 46.14 & 71.45 & 50.87 \\
\bottomrule
\end{tabular}
\caption{Comparison of validation accuracy, Neural Collapse metrics, and OOD generalization (measured via linear probe accuracy) across multiple benchmarks. \texttt{NCAL} consistently achieves stronger generalization to diverse OOD datasets compared to baselines.}
\label{tab:ood_nc_results}
\end{table*}

\paragraph{Covariate Shift Results.}
We test the ability to generalize to OOD datasets by training a linear probe over the learned embeddings. Table \ref{tab:ood_nc_results} shows that NCAL-R improves OOD classification by $\sim 2\%$ on average across 8 varying datasets, over all baselines. This demonstrates the adaptability of NCAL-R's feature space to both NearOOD and FarOOD scenarios.

\paragraph{Label Drift and GCD.}
NCAL-R's geometry-aware selection yields features that support unsupervised novel-class discovery while maintaining high accuracy on known classes. In the GCD setting with 60-40 split, NCAL-R improves novel-class accuracy by $+2.1$ points over the best baseline without supervision on novel classes, and by $+1.6$ points on known classes. This demonstrates that NCAL-R's feature space is inherently adaptable to evolving label spaces, without forgetting past label information.

\begin{figure*}[ht]
    \centering
    \begin{subfigure}[t]{0.32\textwidth}
        \centering
        \includegraphics[width=\linewidth]{raw_images/histogram_zoomed.png}
        \caption{Inter-Class Separation in Feature Space}
        \label{fig:histogram}
    \end{subfigure}
    \hfill
    \begin{subfigure}[t]{0.32\textwidth}
        \centering
        \includegraphics[width=\linewidth]{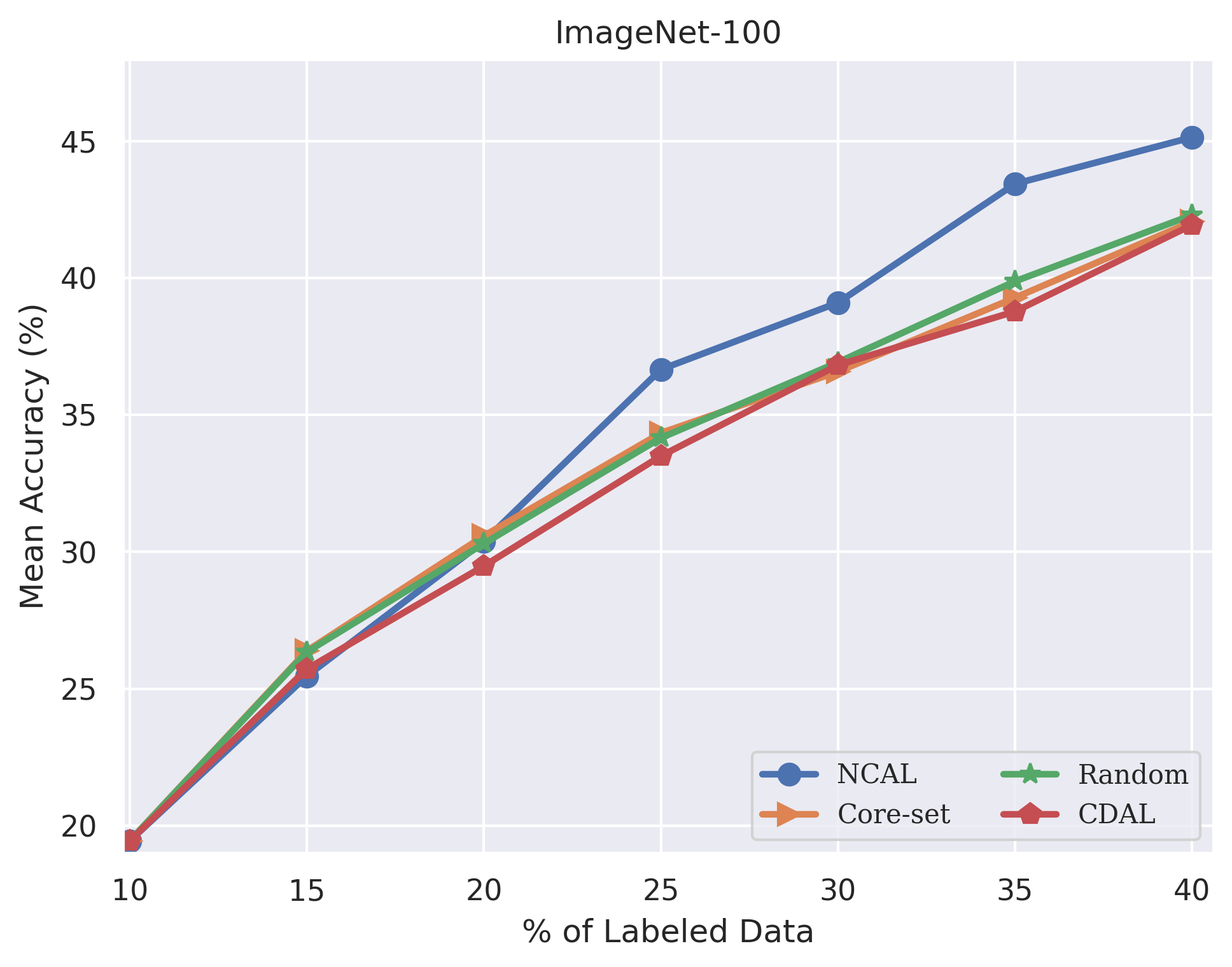}
        \caption{Comparison in Long-Tail Distribution}
        \label{fig:imbalance}
    \end{subfigure}
    \hfill
    \begin{subfigure}[t]{0.32\textwidth}
        \centering
        \includegraphics[width=\linewidth]{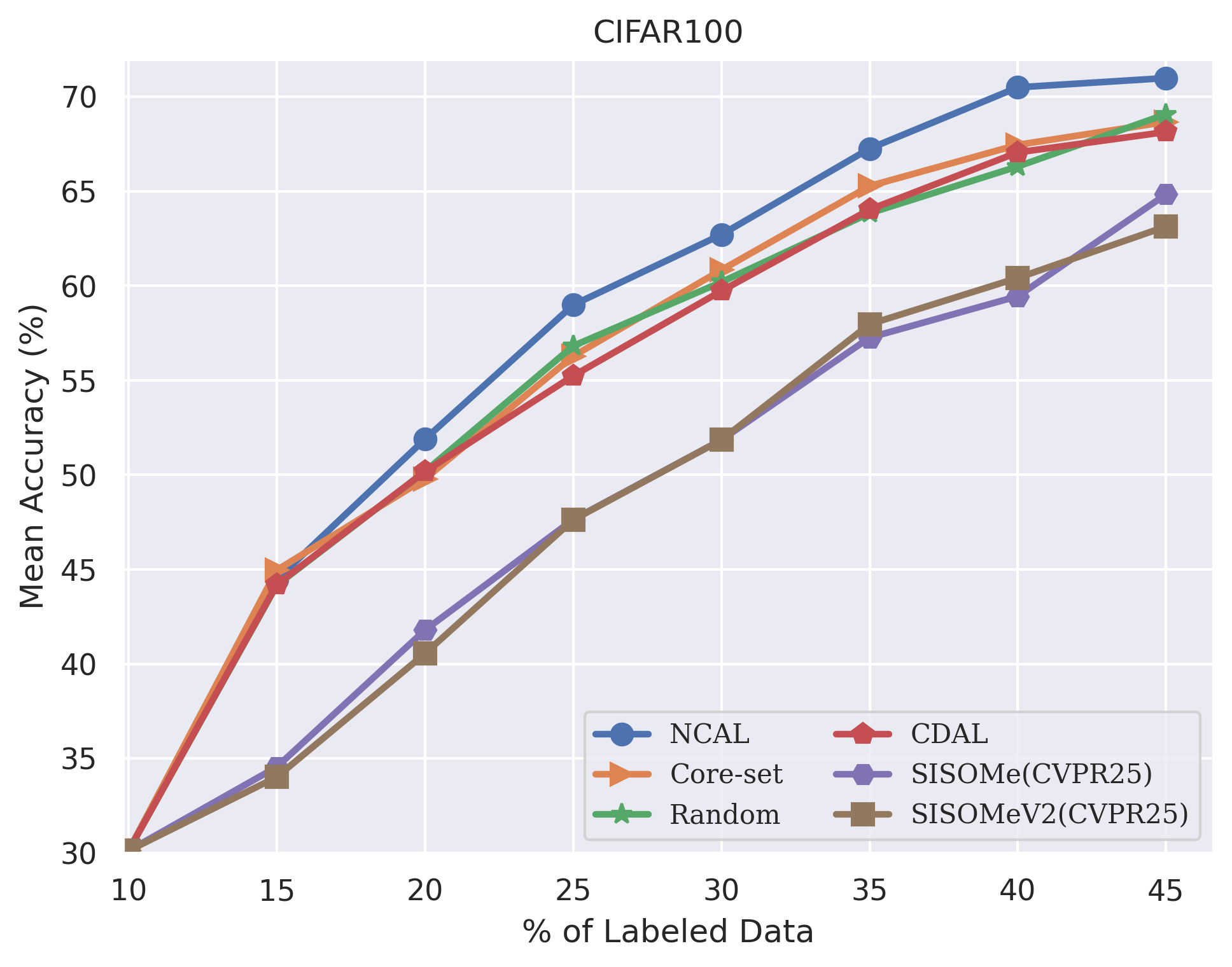}
        \caption{Comparison with ActiveOOD}
        \label{fig:activeOOD_ablation}
    \end{subfigure}
    \caption{Ablation}
    \label{fig:ablation}
\end{figure*}

\mypara{Inter-Class Separation in Feature Space}
To further analyze the structure of learned representations, we examine the distribution of inter-class distances in the penultimate feature space. \cref{fig:histogram} shows a density plot comparing these distributions across different Active Learning strategies. Notably, \texttt{NCAL-R} exhibits a clear rightward shift, indicating larger average separation between class centroids (mean = 15.944), compared to Random (15.114), Coreset (15.070), and CDAL (15.130). This increased inter-class distance suggests that \texttt{NCAL-R} promotes more discriminative and geometrically separated class representations an essential property for improving generalization, especially under low-label regimes and OOD scenarios.

\mypara{Performance Comparison in Long-Tail Distribution}
Real-world data comes in a long-tail distributions, leading to bias towards certain classes. We construct a highly imbalanced version of ImageNet-100 by applying an exponential decay to class sample counts with a decay factor of $\beta=0.05$, leading to a pool of 41,454 samples. An active-learning cycle with this pool achieves 45.15\% for \texttt{NCAL-R}, compared to 42.30\% (Random), 42.06\% (Coreset) and 41.94\% (CDAL) an improvement of +3\% with only 16k images \cref{fig:imbalance}.

\mypara{Evaluating Transferability of ActiveOOD Strategies}
In this ablation, we evaluate the recently proposed ActiveOOD technique SISOMe \cite{schmidt2025joint} for Open-Set in our Closed-Set AL setup by removing its OOD filtering component. As shown in \cref{fig:activeOOD_ablation}, SISOMe performs significantly worse than both standard baselines and NCAL. These results indicate that SISOMe’s scoring heuristics do not transfer well to settings without explicit OOD filtering. 

\section{Related Work}
\label{sec:related}


\mypara{Limitations of Active Learning}
Active Learning reduces labeling cost by prioritizing informative samples, with strategies typically grouped into uncertainty-based \cite{yoo2019learning, gal2017deep}, diversity-based \cite{sener2018active, cdal}, and representativeness-based methods \cite{sinha2019variational}. Despite advances, several critical limitations remain: (i) limited transferability across datasets and architectures, often requiring task-specific tuning \cite{yoo2019learning, ash2020deep}; (ii) unstable performance across cycles, sensitive to initialization and acquisition frequency \cite{beluch2018power, sinha2019variational}; and (iii) poor scalability as many approaches demand costly computations \cite{gal2017deep, sener2018active}. These methods depend on static heuristics, overlooking the rich geometric structures emerging during training -- a gap that motivates our use of \textit{Neural Collapse} as a signal for AL.




\mypara{Active Learning Beyond In-Distribution}
To improve robustness beyond the labeled in-distribution, Active OOD methods filter unlabeled OOD samples during acquisition using auxiliary classifiers, thresholds, or similarity scores (e.g., JODA \cite{schmidt2025joint}). Such techniques are multi-stage and require careful tuning, which limits scalability. Meanwhile, Active Generalized Category Discovery (Active GCD) aims to identify novel classes in the unlabeled pool through clustering and selective labeling \cite{ma2024active}, but often relies on pseudo-labels, merging heuristics, or assumptions about cluster purity. In contrast, our proposed method takes a unified approach to both OOD generalization and category discovery, without requiring OOD filtering or additional supervision mechanisms.

\mypara{Neural Collapse and Structured Feature Representations}
Neural Collapse (\texttt{NC}) refers to a set of emergent geometric properties discovered in \cite{papyan2020prevalence} during the terminal phase of training (TPT) deep neural networks, when the training error approaches zero and the data is balanced. The learned representation and classifier exhibit simple structure: (i) intra-class features collapse to their mean (\NC\texttt{1}); (ii) class means form a simplex equiangular tight frame (\NC\texttt{2}); (iii) classifier weights align with class means (\NC\texttt{3}); and (iv) prediction reduces to nearest-mean classification (\NC\texttt{4}). This phenomenon suggests that \texttt{NC} provides a lens into how class relationships are encoded in both the feature space and the classification layer. Yet, its potential in \textit{Active Learning} remains largely unexplored.




\section{Conclusion}

We presented \texttt{NCAL-R}, an Active Learning framework that leverages Neural Collapse geometry. By combining CMAP and FF scores, NCAL selects structurally informative and uncertain samples, yielding more discriminative and robust feature spaces. Experiments show consistent gains across accuracy, OOD detection, OOD generalization, and category discovery. At its core, \texttt{NCAL-R} shows that structure matters -- aligning acquisition decisions with the emergent geometry of deep networks can pay significant dividends.

\bibliographystyle{plainnat}
\bibliography{bibliography}

\appendix

\section{Deriving the CMAP}\label{sec:derivation}
In this section, we derive the equation for the Class-Mean Alignment Perturbation (CMAP) score, as introduced in Sec. \ref{sec:method}. The CMAP quantifies the change in alignment of the normalized class means induced by candidate sample $x \in \mathcal{U}_t$.


Let's define the Class-Mean Alignment (CMA) before and after adding the sample, respectively. This expression is simply the pair-wise average cosine similarity of class means. 
Then,
\begin{align*}
\textrm{CMA}(\mathcal{L}_t) &= \frac{1}{K(K-1)} \sum_{\substack{i,j=1 \\ i \neq j}}^{K} \textrm{Sim}(\mu_t^i, \mu_t^j) \\
&= \frac{1}{K(K-1)} \left[
\sum_{\substack{i,j=1 \\ i \neq j \\ i \neq c \\ j \neq c}} \text{Sim}(\mu_t^i, \mu_t^j)
+ 2 \sum_{\substack{i=1 \\ i \neq c}}^{K} \text{Sim}(\mu_t^i, \mu_t^c)
\right]
\end{align*}

Given $\mathcal{L}_t \cup x$, and predicted label $\hat{y}(x) = c$, the embedding mean for class $c$ is updated to $\tilde{\mu}_t^c = \frac{|\mathcal{L}_t^c|\mu_t^c + z}{|\mathcal{L}_t^c| + 1}$, while all other means remain unchanged. In the above expression, we isolated the terms involving class $c$ since only those are affected by the perturbation. The remaining terms cancel out when computing the delta:
\begin{align*}
        \mathrm{CMAP}(x) &:= \mathrm{CMA}(\mathcal{L}_t \cup x) - \mathrm{CMA}(\mathcal{L}_t) \nonumber \\
&= \frac{2}{K(K-1)} \sum_{\substack{i=1 \\ i \neq c}}^{K} \left[
\text{Sim}(\mu_t^i, \tilde{\mu}_t^c) - \text{Sim}(\mu_t^i, \mu_t^c)
\right]
\end{align*}

Using cosine similarity, $\text{Sim}(a, b) = \frac{a^T b}{\|a\| \|b\|}$, and denoting $\bar{\mu} = \frac{\mu}{\|\mu\|}$ as the unit-norm version of a vector, we simplify the expression:
\begin{align*}
\textrm{CMAP}(x) &= \frac{2}{K(K-1)} 
\sum_{\substack{i=1 \\ i \neq c}}^{K} 
\left[
\left( \bar{\tilde{\mu}}_t^c \right)^{\!T} \bar{\mu}_t^i 
- 
\left( \bar{\mu}_t^c \right)^{\!T} \bar{\mu}_t^i
\right] \\
&= \frac{2}{K(K-1)} 
\sum_{\substack{i=1 \\ i \neq c}}^{K} 
\left[
\left( \bar{\tilde{\mu}}_t^c - \bar{\mu}_t^c \right)^{\!T} \bar{\mu}_t^i
\right] \\
&= \frac{2}{K(K-1)} 
\left( \bar{\tilde{\mu}}_t^c - \bar{\mu}_t^c \right)^{\!T}
\sum_{\substack{i=1 \\ i \neq c}}^{K} \bar{\mu}_t^i \\
&= \frac{2}{K(K-1)} 
\left( \bar{\tilde{\mu}}_t^c - \bar{\mu}_t^c \right)^{\!T}
\left( M_t - \bar{\mu}_t^c \right)
\end{align*}

Finally, omitting the constant for interpretability and ranking purposes, we define the perturbation score:
\begin{align*}
\boxed{
\text{CMAP}(x) = (\bar{\tilde{\mu}}_t^c - {\bar{\mu}_t^c)}^T (M_t - \bar{\mu}_t^c)
}
\end{align*}

\paragraph{Implementation note:} \(\mathrm{CMAP}\) requires only the current per-class counts \(\{|\mathcal{L}_t^c|\}\) and means \(\{\mu_t^c\}\) plus the feature \(z\) for \(x\); the increment \(\tilde{\mu}_t^c\) can be computed cheaply and \(M_t\) updated incrementally if desired.

\section{Training Protocol}\label{sec:training-protocol}

At each AL cycle:
\begin{enumerate}
    \item Train $f_\theta$ on $\mathcal{L}$ until the Neural Collapse phase.
    \item Compute CMAP and FF for all $x \in \mathcal{U}_t$.
    \item Select $\mathcal{B}_t$ using the combined score, query labels, and update $\mathcal{L}_{t+1} \leftarrow \mathcal{L}_t \cup \mathcal{B}_t$.
    \item Repeat until the budget is exhausted.
\end{enumerate}

\paragraph{Experimental settings.} 
\begin{enumerate}
    \item \textbf{ImageNet100:} Initial pool consists of 10\% randomly sampled data, i.e. 13,000 samples. In each iteration, we select 5\% (i.e., 6,500) samples to be annotated and added to the pool for next iteration of training. We terminate the loop when our labelled pool reaches 40\% of the training set.
    \item \textbf{CIFAR100:} 
    The initial pool size is 10\%, i.e. 5,000 images, acquiring 5\% (2,500 images) in each cycle. We terminate at 45\% pool size.
    \item \textbf{CIFAR10:} The initial pool is 2\% (i.e. 1,000 images), acquiring 2\% images every cycle until 20\% pool size.
\end{enumerate}

\paragraph{Compute.} We run all our experiments on an A100 GPU with a 20 GB memory capacity.

\section{Algorithm Pseudo Code}\label{sec:pseudocode}
\begin{algorithm}
\caption{NCAL Acquisition Function}
\begin{algorithmic}[1]
\STATE \textbf{Input:} Unlabeled pool $\mathcal{U}_t$, labeled set $\mathcal{L}_t$, class means $\{\mu_t^c\}$, model checkpoints $\{f_{\theta_t}\}_{t=T_i}^{T_f}$, acquisition budget $k$
\STATE \textbf{Output:} Selected sample indices $\mathcal{A} \subset \mathcal{U}_t, |\mathcal{A}| = k$

\STATE Initialize empty lists $\{\delta_x\}$ and $\{\phi_x\}$ of size $|\mathcal{U}_t|$ for CMAP and FF respectively
\STATE Compute normalized class means $\bar{\mu}_t^c = \frac{\mu_t^c} {\|\mu_t^c\|}$ for each class $c$
\STATE Compute $M_t := \sum_{c} \bar{\mu}_t^c$

\FORALL{$x \in \mathcal{U}_t$}
    \STATE $c \leftarrow \hat{y}_t(x)$ \COMMENT{Predicted label for $x$}
    \STATE $z \leftarrow$ penultimate-layer feature of $x$
    \STATE $\tilde{\mu}_t^c \leftarrow \frac{|\mathcal{L}_t^c|\mu_t^c+z}{|\mathcal{L}_t^c|+1}$
    \STATE $\bar{\tilde{\mu}}_t^c \leftarrow \tilde{\mu}_t^c / \|\tilde{\mu}_t^c\|$
    \STATE $\delta_x \leftarrow (\bar{\tilde{\mu}}_t^c - {\bar{\mu}_t^c)}^T(M_t - \bar{\mu}_t^c)$

    \STATE $\phi_x \leftarrow 0$
    \FOR{$t = T_i + 1$ \TO $T_f$}
        \IF{$f_t(x) \ne f_{t-1}(x)$}
            \STATE $\phi_x \leftarrow \phi_x + 1$
        \ENDIF
    \ENDFOR
\ENDFOR

\STATE Standardize scores using Z-score normalization:
\[
\widehat{\mathrm{CMAP}}(x) \leftarrow \frac{\delta_x - \mu_\delta}{\sigma_\delta}, \quad
\widehat{\mathrm{FF}}(x) \leftarrow \frac{\phi_x - \mu_\phi}{\sigma_\phi}
\]

\STATE Compute acquisition scores: $s_x := \frac{\widehat{\mathrm{CMAP}}(x) + \widehat{\mathrm{FF}}(x)}{2}$ for each $x \in \mathcal{U}_t$

\STATE Select top-$k$ samples: $\mathcal{A} \leftarrow \texttt{TopK}(\{s_x\}_{x \in \mathcal{U}_t}, k)$

\RETURN $\mathcal{A}$

\end{algorithmic}
\end{algorithm}

\section{Limitations of NCAL-R}\label{sec:limitations}

\paragraph{Limitations.} 
NCAL-R relies on models being trained into the \emph{neural collapse} regime, i.e., the terminal phase where training accuracy plateaus and geometric regularities emerge. Reaching this phase can require many epochs, depending on the dataset and architecture, which may limit efficiency. Moreover, the study of Neural Collapse in large-scale models (e.g., LLMs) remains limited. Since such models are typically trained for only a few epochs, it is unclear whether NCAL-R’s assumptions hold in these settings. We have not evaluated NCAL-R under such large-scale regimes, and adapting it there may require further investigation.

\end{document}